\documentclass[letterpaper, 10 pt, conference]{ieeeconf}  

\IEEEoverridecommandlockouts                              

\usepackage{graphics} 
\usepackage{epsfig} 
\usepackage{mathptmx} 
\usepackage{times} 
\usepackage{amsmath} 
\usepackage{amssymb}  
\usepackage[T1]{fontenc}
\usepackage{multicol}
\usepackage{float}
\usepackage{balance}
\usepackage{indentfirst}
\usepackage{bm}
\usepackage{cite}
\usepackage{subfigure}
\usepackage[ruled]{algorithm2e}
\usepackage{setspace}
\usepackage{algpseudocode}
\usepackage{color}

\title{\LARGE \bf
DOB-Net: Actively Rejecting Unknown Excessive Time-Varying Disturbances
}

\author{Tianming Wang$^{1}$, Wenjie Lu$^{1}$, Zheng Yan$^{2}$ and Dikai Liu$^{1}$
\thanks{This work was supported in part by the Australian Research Council Linkage Project (LP150100935), the Roads and Maritime Services of NSW, and the Centre for Autonomous Systems at the University of Technology Sydney.}
\thanks{$^{1}$Tianming Wang, Wenjie Lu and Dikai Liu are with Centre for Autonomous Systems (CAS), Faculty of Engineering and Information Technology (FEIT), University of Technology Sydney (UTS), 
	Ultimo, NSW 2007, Australia 
	{\tt\small tianming.wang@student.uts.edu.au \{wenjie.lu,dikai.liu\}@uts.edu.au}}%
\thanks{$^{2}$Zheng Yan is with Centre for Artificial Intelligence (CAI), FEIT, UTS, 
	Ultimo, NSW 2007, Australia
	{\tt\small yan.zheng@uts.edu.au}}%
}

\begin{document}

\maketitle
\thispagestyle{empty}
\pagestyle{empty}

\begin{abstract}
	
This paper presents an observer-integrated Reinforcement Learning (RL) approach, called Disturbance OBserver Network (DOB-Net), for robots operating in environments where disturbances are unknown and time-varying, and may frequently exceed robot control capabilities.
The DOB-Net integrates a disturbance dynamics observer network and a controller network. Originated from conventional DOB mechanisms, the observer is built and enhanced via Recurrent Neural Networks (RNNs), encoding estimation of past values and prediction of future values of unknown disturbances in RNN hidden state. Such encoding allows the controller generate optimal control signals to actively reject disturbances, under the constraints of robot control capabilities. The observer and the controller are jointly learned within policy optimization by advantage actor critic. 
Numerical simulations on position regulation tasks have demonstrated that the proposed DOB-Net significantly outperforms a conventional feedback controller and classical RL algorithms.

\end{abstract}

\section{INTRODUCTION}


Autonomous Underwater Vehicles (AUVs) have become vital assets in search and recovery, exploration, surveillance, monitoring, and military applications \cite{griffiths2002technology}. For large AUVs in deep water applications, the strength and changes of external wave and current disturbances are negligible to the AUVs, due to their considerable size and thrust capabilities. While small AUVs are required for some shallow water applications, like bridge pile inspection \cite{woolfrey2016kinematic}, where the disturbances coming from the turbulent flows may frequently exceed AUVs' thrust capabilities. These unknown disturbances inevitably bring adverse effects and may even destabilize robots \cite{xie2000much}. 
Thus this paper studies an optimal control problem of robots subject to excessive time-varying disturbances, and presents an observer-integrated RL solution. 
Such problem also arises in many other applications, e.g., aerial quadrotors for surveillance in wind conditions \cite{waslander2009wind} and manipulators operating with constantly varying loads. In the cases of robot's actuator failure, the control capabilities may also be lower than the external disturbances. 

\begin{figure}[t]
	\centering
	\includegraphics[width=0.7\hsize]{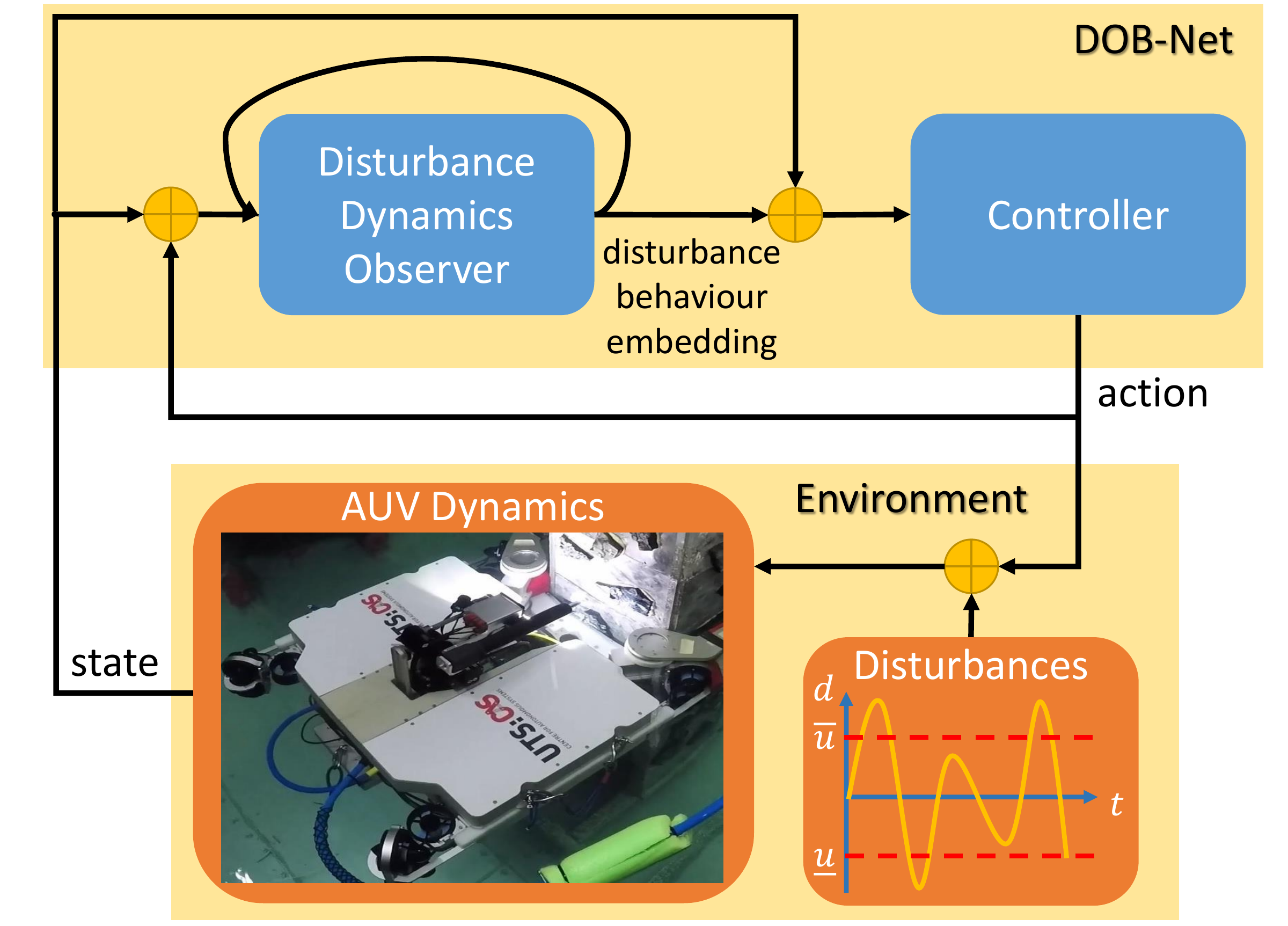}
	\caption{Working flow of the DOB-Net ($\overline{u}$ and $\underline{u}$ are control limits).}
	\label{algorithm_diagram}
\end{figure}


RL \cite{sutton2018reinforcement} is a trial-and-error method that does not require an explicitly system model, and can naturally adapt to noises and uncertainties in the real system. With recent advances in deep neural network, RL is now able to solve practical problems. 
However, the excessive disturbances are not appropriate to be regarded as noises any more, since AUV's state transition is heavily affected by the external disturbances, thus violating the assumption of Markov Decision Process (MDP).
While considering the time-varying characteristics of the current and wave disturbances, if future disturbances can be predicted, RL may be able to generate optimal controls.

Conventional DOB \cite{chen2000nonlinear} and related methods have been researched and applied in various industrial sectors in the last four decades. The main objective of the use of DOB is to deduce the unknown disturbances from measurable variables, without additional sensors. Then, a control action can be taken, based on the disturbance estimate, to compensate for the influence of the disturbances, which is called Disturbance-Observer-Based Control (DOBC) \cite{chen2016disturbance}.

However, conventional DOBC does have some limitations when solving our problem. The first limitation is that DOB normally needs an accurate system model, which is usually unavailable for underwater vehicles due to complex hydrodynamics. In this case, model uncertainties are lumped with external disturbances, and then estimated by DOB together. Thus, the original properties for some disturbances, such as harmonic ones, are changed.
%
%
Furthermore, even with an sufficiently accurate estimate of the disturbances at current timestep, the optimal control solution is still unreachable due to the neglection of time correlation of the disturbance signals. The reason behind this is that disturbances exceeding control constraints cannot be well rejected only through feedback regulation.
Sun and Guo \cite{sun2016neural} proposed a neural network approach to construct DOBC. As compared with conventional DOBC methods, the primary merit of the proposed method is that the model uncertainties are approximated using a radial basis function neural network (RBFNN) technique, and not regarded as part of the disturbances. Then the external disturbances can be separately estimated by a conventional DOB. However, they still failed to consider the control constraints, leading to poor control capabilities under excessive disturbances.
In order for optimal overall performance, the AUV behaviour needs to be optimized over a future time horizon using a sequence of disturbance estimates. 


This paper proposes a novel RL approach called DOB-Net, which enables integrated learning of disturbance dynamics and an optimal controller, for current and wave disturbance rejection control of AUVs in shallow and turbulent water, as shown in Fig~\ref{algorithm_diagram}. 
The DOB-Net consists of a disturbance dynamics observer network and a controller network. The observer network is built and enhanced via RNNs, through imitating the conventional DOB mechanisms. But this network is more flexible than the conventional one, since it encodes the prediction of the external disturbances in RNN hidden state, instead of only estimating the current value of disturbances. Also, the observer function is more robust to the model uncertainties and the rapidly time-varying characteristics of the external disturbances.
Based on the encoded disturbance prediction, the controller network is able to actively reject the unknown disturbances. The observer and the controller are jointly learned within policy optimization by advantage actor critic. This integrated learning may achieve an optimized representation of observer outputs, compared with traditional hand-designed features.
The policy is trained using simulated disturbances consisting of multiple sinusoidal waves, and evaluated using both simulated disturbances and collected disturbances, the latter is gathered from an Inertial Measurement Unit (IMU) onboard an AUV in a water tank at University of Technology Sydney. During training, the amplitude, period and phase of sine-wave disturbances are randomly sampled in each episode.


In this paper, related work is presented in Section II. Section III introduces problem formulation. Section IV provides the detailed description of the DOB-Net. Then, Section V presents validation procedures and result analysis. Some potential future improvements are discussed in the last section.

\section{RELATED WORK}

\subsection{Feedback and Predictive Control}

In the early development of disturbance rejection control, feedback control strategies are used to suppress the unknown disturbances \cite{edwards1998sliding,lu2017active,lu2018frequency}. 
Then, disturbance estimation and attenuation methods through adding a feedforward compensation term have been proposed and practiced, such as DOBC \cite{chen2016disturbance,chen2000nonlinear}.
However, these methods often assume that the system deals with bounded disturbances which should be small enough, thus fail to guarantee stability considering control constraints when meeting strong disturbances \cite{gao2016nonlinear}.

%
%

To this end, Model Predictive Control (MPC) \cite{garcia1989model} is often applied due to its constraint handling capacity through optimizing plant behaviour over a certain time horizon \cite{gao2016nonlinear}. The MPC requires a prediction model of the system to optimize future behaviour, this model includes not only the robot dynamics, but also the predicted disturbances over next optimization horizon. Thus, researchers have developed a compound control scheme consisting of a feedforward compensation part based on the conventional DOB and a feedback regulation part based on the MPC (DOB-MPC) \cite{maeder2010offset}. 
However, the performance of the MPC heavily relies on the accuracy of given system model, and the requirement for online optimization at each timestep leads to a low computational efficiency. Besides, such separated modeling and control optimization process might not be able to produce models and controls that jointly optimize robot performance, as evidenced in \cite{brahmbhatt2017deepnav}. In contrast, the DOB-Net uses neural networks to construct both the observer and the controller, achieving model-free control, high run-time efficiency as well as a joint optimization of the observer and the controller.

\subsection{Classical RL}

RL has drawn a lot of attention in finding optimal controllers for systems that are difficult to model accurately. Recently, deep RL algorithms based on Q-learning \cite{mnih2015human}, policy gradients \cite{schulman2015trust}, and actor-critic methods \cite{mnih2016asynchronous} have been shown to learn complex skills in high-dimensional state and action spaces, including simulated robotic locomotion, driving, video game playing, and navigation. 
RL generally considers stochastic systems of the form \cite{saemundsson2018meta}
\begin{equation}
x_{t+1}=f\left(x_{t}, u_{t}\right)+\epsilon,
\end{equation}
with state variables $x \in \mathbb{R}^{D}$, control signal $u \in \mathbb{R}^{K}$ and i.i.d. system noise marginalized over time $\epsilon \sim \mathcal{N}(0, E)$, where $E = \operatorname{diag}\left(\sigma_{1}^{2}, \ldots, \sigma_{D}^{2}\right)$. While in our case, the current and wave disturbances should be regarded as functions of time instead of random noises, refer to \eqref{dynamics}, due to its large amplitudes and time-varying characteristics, as evidence in Section V.

\subsection{History Window Approach}

When using RL to deal with unknown disturbances, the problem cannot be defined as a MDP, since the transition function does not only depend on the current state and action, but also heavily on the disturbances. The history window approaches \cite{lin1993reinforcement} attempt to resolve the hidden state by making the selected action depend not only on the current state, but also on a fixed number of the most recent states and actions. 
Wang et al. \cite{wang2018excessive} applied this approach to handle the external disturbances of an AUV through characterizing the disturbed AUV dynamics model as a multi-order Markov chain $x_{t+1} = f_{h}(H_{t}, x_{t}, u_{t})$, and assuming the unobserved time-varying disturbances and their prediction over next planning horizon are encoded in state-action history of fixed length $H_{t} = \{ x_{t-N}, u_{t-N}, \cdots , x_{t-1}, u_{t-1} \}$, where $N$ represents the history length. Then, the policy is trained to take a fixed length of state-action history along with the current state as input to generate control signals. 
%
However, it is difficult to determine an optimal length of the history. Shorter history may not provide sufficient information about disturbances, longer history will make the training difficult. Wang et al. \cite{wang2018excessive} considered the history length as a hyperparameter that was statistically optimized during training.


\subsection{Recurrent Policy}

Due to the difficulty in determining optimal history length through history window approach, RNN is then utilized to automatically learn how much past experience should be explored to achieve optimal performance.
%
Using RNN to represent policies is a popular approach to handle partial observability \cite{hausknecht2015deep,wierstra2007solving}. The idea being that the RNN will be able to retain information from states further back in time and incorporate this information into predicting better actions and value functions and thus performing better on tasks that require long term planning.
%
Particularly, a RNN policy produces an action and a hidden vector based on current state and a hidden vector from last timestep. Since the hidden vector contains past processed information, the action is a function that depends on all the previous states.
These policies are able to solve tasks that require memory by loading sequence of states \cite{wierstra2007solving}. However, most of them only considered the state-only history, which, for example, has been used for estimating velocities in training video game player \cite{mnih2015human}. 

Sutskever et al. \cite{sutskever2014sequence} presented a novel usage of RNN, where the input to the RNN comes from the previously predicted output. This architecture gives rise to an idea of using both past states and actions in the recurrent RL approach, where the recurrent policy produces an action $u_{t}$, given both current state $x_{t}$ as well as previously executed action $u_{t-1}$ as the RNN input. These approaches encode additional dynamics information besides kinematics information, enabling the capability of observing disturbances. Compared with the recurrent RL using state-action history, the contributions of DOB-Net lie in the exploration and application of the architectural similarities between Gated Recurrent Unit (GRU) and conventional DOB, and then the ability to encode the prediction of disturbance dynamics function.


\subsection{Meta-Reinforcement Learning}


Meta-Reinforcement Learning (Meta-RL) \cite{finn2017model,nagabandi2018learning,rakelly2019efficient} defines a framework which leverages data from previous tasks to acquire a learning procedure that can quickly adapt to new tasks using only a small amount of experience. The tasks are considered as MDPs or Partially Observable MDPs (POMDPs) drawn from a task distribution with varying transition functions or varying reward functions. Nagabandi et al. \cite{nagabandi2018learning} and Rakelly et al. \cite{rakelly2019efficient} both considered adaptation to the current task setting using past transitions. Rakelly et al. \cite{rakelly2019efficient} considered adaptation at test time in meta-RL as a special case of RL in a POMDP by including the task as the unobserved part of the state, and learned a probabilistic latent representation of prior experience to capture uncertainty over the task. 
%

However, these papers either assume each task is an invariable MDP \cite{finn2017model,rakelly2019efficient}, or assume the environment is locally consistent during a rollout \cite{nagabandi2018learning}, which may not be suitable for our problem due to the rapid time-varying characteristic of the current and wave disturbances. Also, these disturbances are better described as functions of time, instead of just uncertainties in the transition function. In that case, the external disturbances (or variations in transition function) for two consecutive samples are not independent and identically distributed (i.i.d.). Current meta-RL formulation has not clearly indicated an appropriate solution for this time correlated variations in the transition function. A potential idea is to characterize the disturbed transition function as a multi-step MDP, where the state space contains not only the current state, but also the most recent states and actions. This naturally leads us to the history window approach and recurrent RL.

\section{PROBLEM FORMULATION}

\subsection{System Description}

Our 6 Degree Of Freedom (DOF) AUV is designed to be sufficiently stable in orientation even under strong disturbances, thanks to its large restoring forces. Thus, we only consider the disturbance rejection control of the vehicle’s 3-DOF position, and assume its orientation is well controlled all the time. However, the framework can be easily extended to 6-DOF case, where a larger network and longer training time may be required. And it is also applicable for other kinds of mobile robots subject to excessive time-varying disturbances, such as quadrotors, gliders, and surface vessels, but the effectiveness needs further investigation.

The AUV model can be considered as a floating rigid body with external disturbances, which can be represented by
\begin{equation}\label{dynamics}
\begin{array}{c}
{M(q) \ddot{q}+G(q, \dot{q})=u+d(t)} \\
{G(q, \dot{q})=C(q, \dot{q}) \dot{q}+D(q, \dot{q}) \dot{q}+g(q)} 
\end{array},
\end{equation}
where $M(q) \in \mathbb{R}^{3 \times 3}$ is the inertia matrix, $C(q,\dot{q})  \in \mathbb{R}^{3 \times 3}$ is the matrix of Coriolis and centripetal terms, $D(q,\dot{q})  \in \mathbb{R}^{3 \times 3}$ is the matrix of drag force, $g(q)  \in \mathbb{R}^{3}$ is the vector of the gravity and buoyancy forces, $q,\dot{q},\ddot{q} \in \mathbb{R}^{3}$ represent replacements, velocities and accelerations of the AUV, $u \in \mathbb{R}^{3}$ represents the control forces, $d(t) \in \mathbb{R}^{3}$ is the time-varying disturbance forces, and the variation of $d(t)$ with time from the past to the future is the disturbance dynamics, which is exactly what the observer network tries to produce. The AUV dynamic model is assumed to have fixed parameters, the model and the parameters are not known. In our case, we assume that the magnitudes of the disturbances will exceed the AUV control limits $\overline{u} \in \mathbb{R}^{3}$ and $\underline{u} \in \mathbb{R}^{3}$, but are constrained within reasonable ranges, ensuring the controller is able to stabilize the AUV in a sufficiently small region.


\subsection{Problem Definition}

In RL, the goal is to learn a policy that chooses actions $u_{t}$ at each timestep $t$ in response to the current state $x_{t}$, such that the total expected sum of discounted rewards is maximized over all time. 
The state of the robot consists of position as well as the corresponding velocities $x = [q^{T} \ \dot{q}^{T}]^{T} \in \mathcal{X} \in \mathbb{R}^{6}$. The action includes the control forces $u \in \mathcal{U} \in \mathbb{R}^{3}$.
At each timestep, the system transitions from $x_{t}$ to $x_{t+1}$ in response to the chosen action $u_{t}$ and the transition dynamics function $f : \mathcal{X} \times \mathcal{U} \rightarrow \mathcal{X}$, collecting a reward $r_{t}$ according to running reward function or final reward function 
\begin{equation}
\begin{aligned}
r\left(x_{t}, u_{t}\right)=x_{t}^{T} Q x_{t}+u_{t}^{T} R u_{t}, \quad r_{f}\left(x_{t}\right)=x_{t}^{T} Q x_{t},
\end{aligned}
\end{equation}
where $Q \in \mathbb{R}^{6 \times 6}$ and $R \in \mathbb{R}^{3 \times 3}$ represent weight matrices. The discounted sum of future rewards is then defined as $\sum_{t'=t}^{T-1} \gamma^{t'-t}r\left(x_{t'}, u_{t'}\right) + \gamma^{T-t}r_{f}\left(x_{T}\right)$, where $\gamma \in [0, 1]$ is a discount factor that prioritizes near-term rewards over distant rewards \cite{nagabandi2018neural}.

\section{METHODOLOGY}

The underwater disturbances present great challenges for stabilization control due to its excessive amplitudes as well as rapidly time-varying characteristics.
In this section, a conventional DOB is first compared with a GRU, the results show some similarities in the structure of processing hidden information. Thus, an enhanced observer network for excessive time-varying disturbances is designed using GRUs, encoding the disturbance dynamics into GRU hidden state. A following controller network is then built upon this encoding in order to generate optimal controls.

\subsection{Conventional DOB}

%

\begin{figure}[b]
	\centering
	\subfigure[DOB]{
		\includegraphics[width=0.45\linewidth]{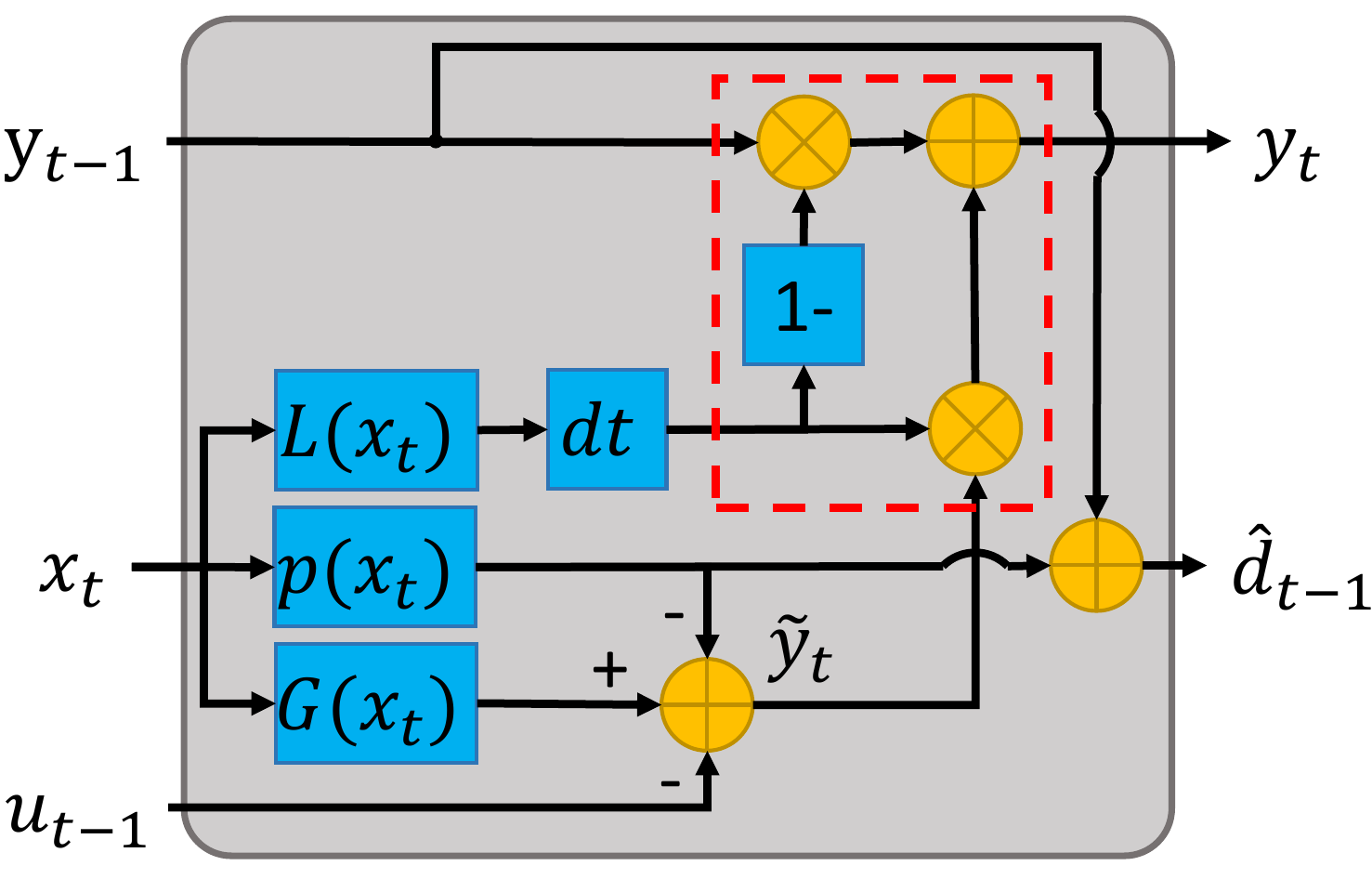}
		\label{fig:architecture_dob}
	}
	\subfigure[GRU]{
		\includegraphics[width=0.45\linewidth]{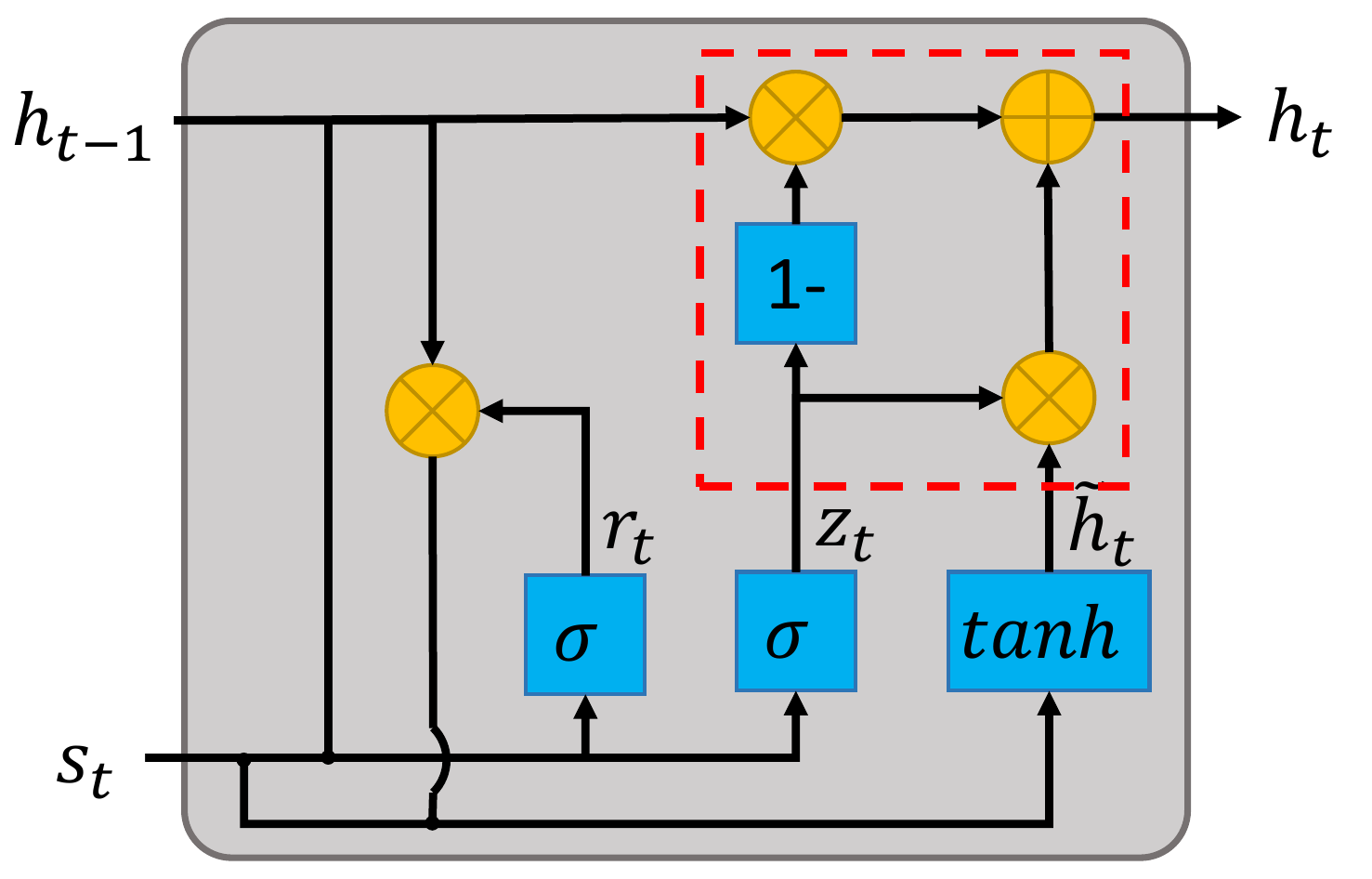}
		\label{fig:architecture_gru}
	}
	\caption{Architecture of DOB and GRU.}
\end{figure}

The basic idea of conventional DOB is to estimate current disturbance forces based on robot state and executed controls, its formulation is proposed as
\begin{equation}
\begin{array}{c}{\dot{y}=-L(q,\dot{q})y+L(q,\dot{q})\left(G(q,\dot{q})-p(q,\dot{q})-u\right)} \\ {\hat{d}=y+p(q,\dot{q})}\end{array},
\end{equation}
where $\hat{d} \in \mathbb{R}^{3}$ is the estimated disturbances, $y \in \mathbb{R}^{3}$ is the internal state of the nonlinear observer and $p(q,\dot{q})$ is the nonlinear function to be designed. The DOB gain $L(q,\dot{q})$ is determined by the following nonlinear function:
\begin{equation}
L(q, \dot{q}) M(q) \ddot{q}=\left[\frac{\partial p(q, \dot{q})}{\partial q} \quad \frac{\partial p(q, \dot{q})}{\partial \dot{q}}\right] \left[ \begin{array}{c}{\dot{q}} \\ {\ddot{q}}\end{array}\right].
\end{equation}
It has been shown in \cite{chen2000nonlinear} that DOB is globally asymptotically stable by choosing
\begin{equation}
L(q, \dot{q})=\operatorname{diag}\{c, c\},
\end{equation}
where $c>0$. More specifically, the exponential convergence rate can be specified by choosing $c$. The convergence and the performance of the DOB have been established for slowly time-varying disturbances and disturbances with bounded rate in \cite{li2016disturbance}.
A discrete version of DOB is also provided (illustrated in Fig.~\ref{fig:architecture_dob})
\begin{equation}
\begin{array}{l}{\tilde{y}_{t}=G(x_{t})-p(x_{t})-u_{t-1}} \\
{y_{t}=\left(1-L(x_{t})dt\right)y_{t-1}+L(x_{t})dt\tilde{y}_{t}} \\ {\hat{d}_{t-1}=y_{t-1}+p(x_{t})}\end{array}.
\end{equation}

\subsection{Gated Recurrent Unit (GRU)}

The architecture of GRU \cite{cho2014learning} is shown in Fig.~\ref{fig:architecture_gru}. The formulations are given below:
\begin{equation}
\begin{array}{l}{z_{t}=\sigma\left(W_{z}\left[h_{t-1}, s_{t}\right]+b_{z}\right)} \\ {r_{t}=\sigma\left(W_{r}\left[h_{t-1}, s_{t}\right]+b_{r}\right)} \\ {\tilde{h}_{t}=\tanh \left(W_{h}\left[r_{t} \circ h_{t-1}, s_{t}\right]+b_{h}\right)} \\ {h_{t}=\left(1-z_{t}\right) \circ h_{t-1}+z_{t} \circ \tilde{h}_{t}}\end{array},
\end{equation}
where $s_{t}$ is the input vector, $h_{t}$ is the output vector, $z_{t}$ is the update gate vector, $r_{t}$ is the reset gate vector, $W$ and $b$ are the weight matrices and bias vectors, $\sigma$ and $tanh$ are the activation functions (sigmoid function and hyperbolic tangent). The operator $\circ$ denotes the Hadamard product.

\subsection{DOB-Net}


\begin{figure}[b]
	\centering
	\includegraphics[height=0.45\linewidth]{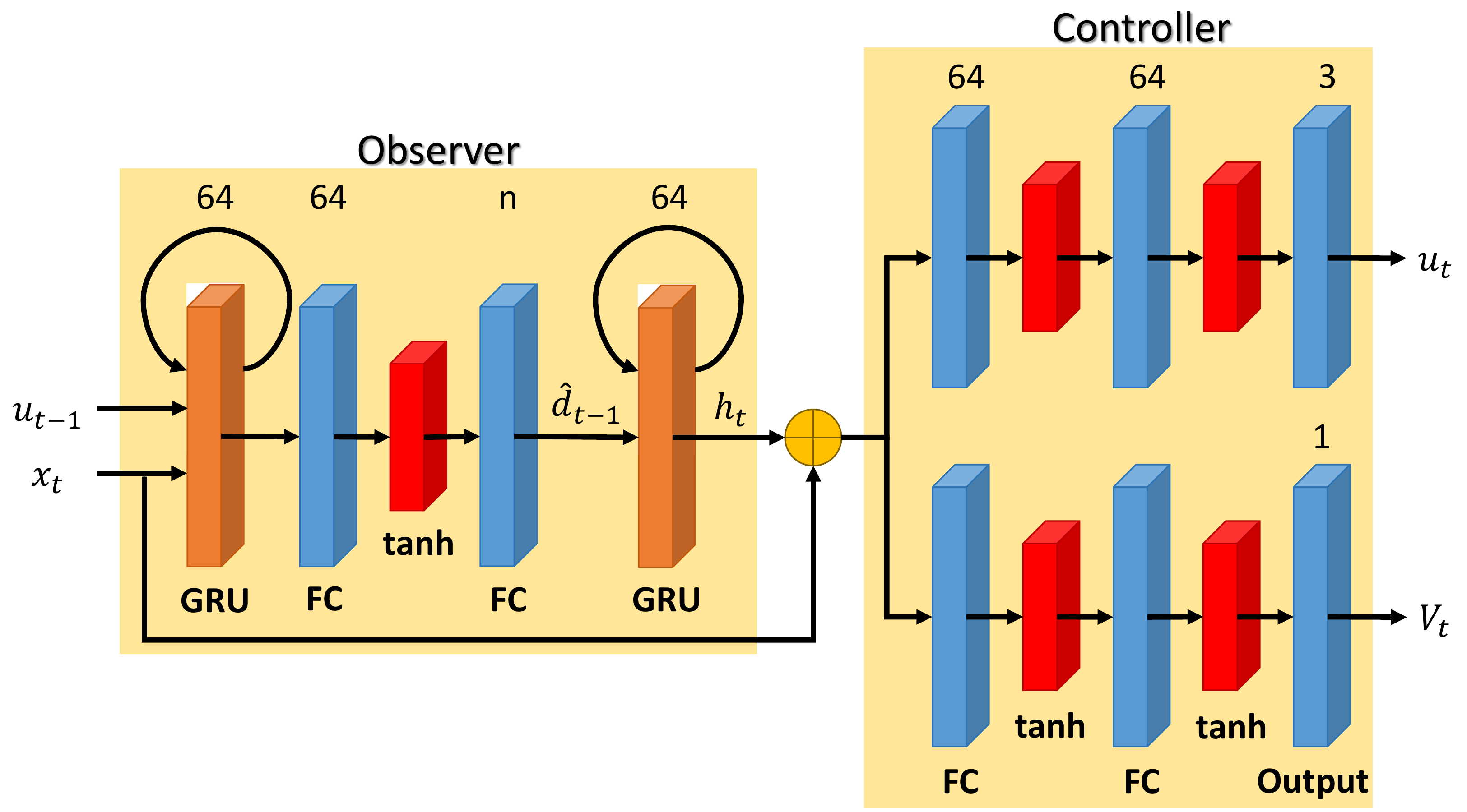}
	\caption{Network architecture of DOB-Net.}
	\label{fig:architecture_dob_net}
\end{figure}

The DOB-Net is constructed based on classical actor-critic architecture.
The observer network consists of two GRUs and two fully connected layers between them, in order to imitate and enhance the function of the conventional DOB. As described in Fig.~\ref{fig:architecture_dob} and Fig.~\ref{fig:architecture_gru}, the DOB and GRU have a similar architecture, especially the part in the red box. $y_{t}$ of DOB acts as the hidden state, similar to the role of $h_{t}$ in GRU, which preserves hidden information for the usage of next timestep. In order to equip GRU with capability of observing disturbances, we first employ a GRU to process the same inputs as DOB, which are the current state $x_{t}$ and the last action $u_{t-1}$. Except for the hidden state, the DOB also outputs the estimated disturbances $\hat{d}_{t-1}$, which is a function of both the input state and the hidden state. Thus, we add fully connected layers after the first GRU to provide better embedding of the disturbance estimation. 

After that, the embedding of the estimated disturbances can be further fed into another GRU, in order to encode a sequence of past disturbances.
The embedding of this disturbance sequence $h_{t}$ is supposed to represent the disturbance dynamics. It can then be combined with the current state $x_{t}$, becoming the actual inputs of the controller network. 
One design parameter of the DOB-Net is the embedding dimension of $\hat{d}_{t-1}$. In this paper, 3-dimension (the size of disturbances) and 64-dimension (the size of RNN hidden state) are chosen and compared in simulation. Such comparison shows the flexibility of neural networks after building observer from GRU.





\begin{algorithm}[t]
	\setstretch{0.92}
	\caption{DOB-Net - pseudocode for each thread} 
	\label{alg:dob_net}	
	Assume global shared parameters $\theta$ and $\theta_{v}$ \\
	Assume thread-specific parameters $\theta^{\prime}$ and $\theta_{v}^{\prime}$ \\
	Initialize global shared counter $T = 0$ \\
	Initialize thread step counter $t = 1$ \\
	\Repeat{$T > T_{\max}$} 
	{
		Reset gradients: $d \theta \leftarrow 0$ and $d \theta_{v} \leftarrow 0$ \\
		Synchronize parameters $\theta^{\prime}=\theta$ and $\theta_{v}^{\prime}=\theta_{v}$ \\
		$t_{start}=t$ \\
		Get state $x_{t}$, last action $u_{t-1}$, last hidden state $h_{t-1}$ \\
		\Repeat{terminal $x_{t}$ or $t-t_{start}==t_{max}$}
		{
			Sample $u_{t}$ according to $\pi\left(u_{t} | x_{t},u_{t-1},h_{t-1} ; \theta^{\prime}\right)$, receive $h_{t}$ \\
			Perform $u_{t}$, receive $r_{t}$ and $x_{t+1}$ \\
			$t \leftarrow t+1$ and $T \leftarrow T+1$ \\
		}
		$R=\left\{\begin{array}{ll}
		{0} & \!\!\!\!\!\! {\text { for terminal } x_{t}} \\ 
		{V\left(x_{t},u_{t-1},h_{t-1} ; \theta_{v}^{\prime}\right)} & \!\!\!\!\!\! {\text { for non-terminal } x_{t}}
		\end{array}\right.$ \\
		\For{$i \in\left\{t-1, \ldots, t_{start}\right\}$}
		{
			$R \leftarrow r_{i}+\gamma R$ \\
			Accumulate gradients wrt $\theta^{\prime} : d \theta \leftarrow d \theta+\nabla_{\theta^{\prime}} \log \pi\left(u_{i} | x_{i},u_{i-1},h_{i-1} ; \theta^{\prime}\right)(R-$ \\
			$\quad V\left(x_{i},u_{i-1},h_{i-1} ; \theta_{v}^{\prime}\right))$ \\
			Accumulate gradients wrt $\theta_{v}^{\prime} : d \theta_{v} \leftarrow d \theta_{v}+\partial\left(R-V\left(x_{i},u_{i-1},h_{i-1} ; \theta_{v}^{\prime}\right)\right)^{2} / \partial \theta_{v}^{\prime}$ \\
		}
		Perform update of $\theta$, $\theta_{v}$ using $d \theta$, $d \theta_{v}$ \\
	}
\end{algorithm}

\textbf{Training:}
Advantage Actor Critic (A2C) \cite{mnih2016asynchronous} is a conceptually simple and lightweight framework for deep RL that uses synchronous gradient descent for optimization of deep neural network controllers.
The algorithm synchronously execute multiple agents in parallel, on multiple instances of the environment. This parallelism also decorrelates the agents’ data into a more stationary process, since at any given time-step the parallel agents will be experiencing a variety of different states.
%

Our algorithm is developed in A2C style. Pseudocode of the DOB-Net is shown in Algorithm~\ref{alg:dob_net}. Each thread interacts with its own copy of the environment. The disturbances are also different in each thread, and each of them are randomly sampled. We found this setting helps accelerate the convergence of learning and improve performance, through comparison with using the same disturbances through all threads during numerical simulations.
The algorithm operates in the forward view by explicitly computing $k$-step returns. 
In order to compute a single update for the policy and the value function, the algorithm first samples and performs actions using its exploration policy for up to $t_{max}$ steps or until a terminal state is reached. The algorithm then computes gradients for $k$-step updates. Each $k$-step update uses the longest possible $k$-step return resulting in a one-step update for the last state, a two-step update for the second last state, and so on for a total of up to $t_{max}$ updates. The accumulated updates are applied in a single gradient step.

\section{SIMULATION EXPERIMENTS}

\subsection{Simulation Setup}

A position regulation task is simulated to test our approaches. The simulated AUV has the mass $m = 60 \ kg$ with the size of $0.8 \times 0.8 \times 0.25 \ m^{3}$. Only positional motion and control are considered, thus the AUV has a 6-dimensional state space and a 3-dimensional action space. The control limits $\left| \overline{u} \right| = \left| \underline{u} \right| = [120N \ 120N \ 120N]^{T}$. 
Each training episode contains 200 steps with 0.05s per step. In each episode, the robot starts at a random position with a random velocity, and it is controlled to reach a target position and stay within a region (refer to as converged region) thereafter. 


%

\begin{figure}[t]
	\centering
	\includegraphics[width=0.9\linewidth]{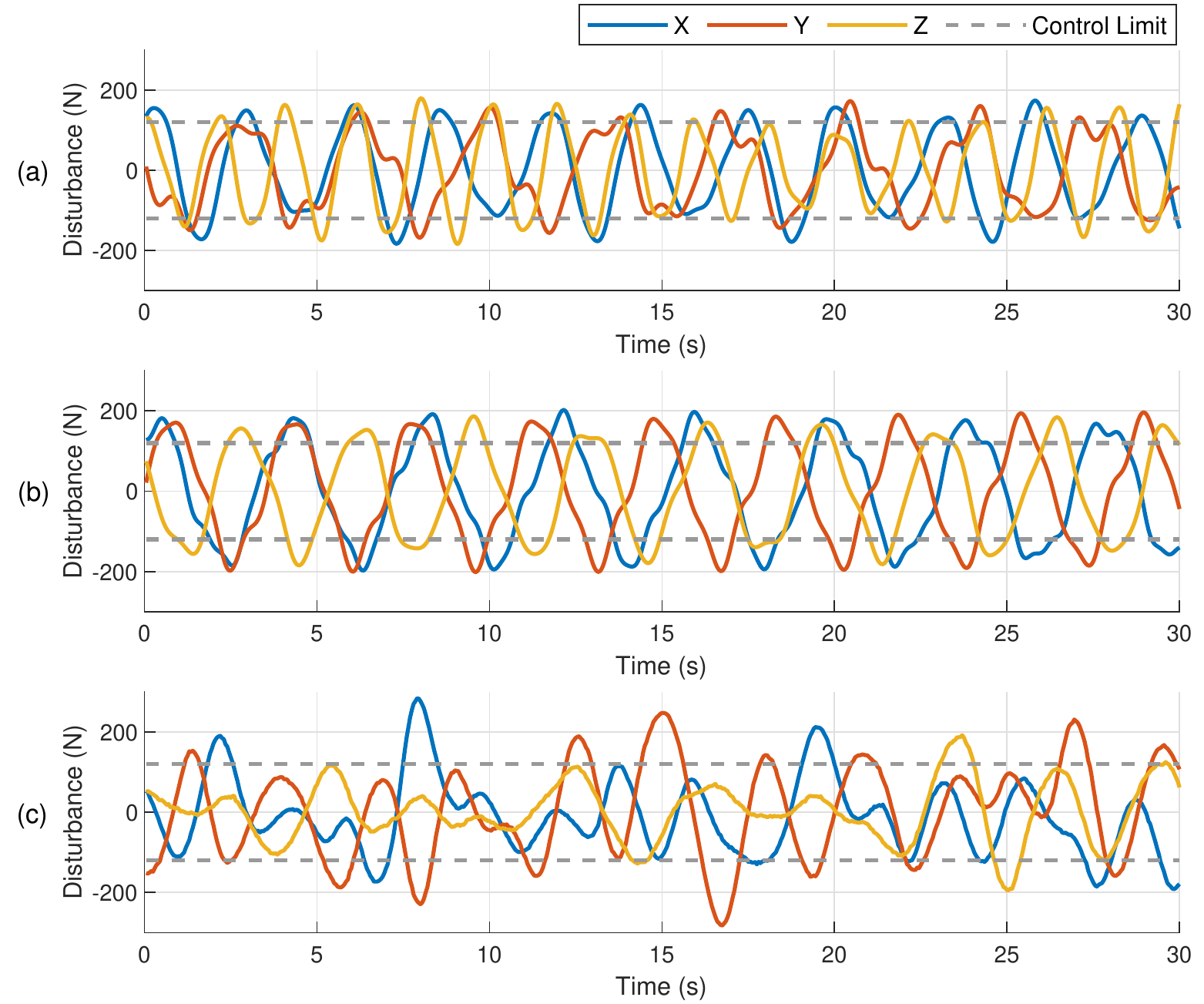}
	\caption{Example disturbances in X, Y and Z directions, (a) small simulated disturbances; (b) large simulated disturbances; (c) collected disturbances.}
	\label{fig:disturbance}
\end{figure}

In these experiments, the algorithms are trained using simulated disturbances, and tested using both simulated disturbances and collected disturbances.
The simulated disturbances are constructed as a superposition of multiple sinusoidal waves (three in our case) with different amplitudes, frequencies and phases. Two different scenarios are considered, one has close or slightly excessive amplitudes (around 100-120\% of control limits), the other one has larger amplitudes (130-150\% of control limits). Examples of the simulated disturbances are given in Fig.~\ref{fig:disturbance} (a) and (b). 
Our purpose is to enable the trained policy to deal with unknown time-varying disturbances, thus their amplitudes, periods and phases are randomly sampled from the given distributions in each training episode.
In order to further validate the efficacy of the proposed algorithms, we also collected the current and wave disturbance data in a water tank using wave generator, as shown in Fig.~\ref{fig:disturbance} (c). The data is collected through an onboard IMU of an unactuated AUV, the measured linear accelerations are mapped to forces, which can be assumed as the external disturbances. 
We notice that the amplitudes of the collected disturbances are not constrained within the ranges seen during training, leading to a more challenging scenario.

%

Ten different approaches for disturbance rejection control are tested and compared:

(a) Robust integral of the sign error (RISE) control \cite{fischer2014nonlinear}

(b) DOBC

(c) A2C

(d) History window A2C with state history (HWA2C-x)

(e) History window A2C with state-action history (HWA2C-xu)

(f) Recurrent A2C with state history (RA2C-x)

(g) Recurrent A2C with state-action history (RA2C-xu)

(h) DOB-Net ($n=3$)

(i) DOB-Net ($n=64$)

(j) Trajectory optimization

\noindent Notice that, Among these methods, the trajectory optimization assumes full knowledge of the disturbances over the whole episode, while all other algorithms deal with unknown disturbances. The comparison is obviously not fair, the trajectory optimization is used only to provide a performance in ideal case.
RISE control is a conventional feedback controller; DOBC is an implementation of Sun and Guo's work \cite{sun2016neural}, considering control constraints; HWA2C applies the history window approach into the A2C framework, the used window length is 10 timesteps, which is 0.5s in our simulation setup; and RA2C employs RNNs to deal with the past states and actions.
The applied A2C framework employs a parallel training mode, 16 agents are used at the same time, the equivalent real-world training time for each agent is 43.4 hours. The RL algorithms accumulate gradients over 20 steps, its reward discount factor is 0.99. The used optimizer for RL is RMSprop, where the learning rate is 7e-4 and the max norm of gradients is 0.5.
In the remaining part of this section, we first evaluate the training process of different algorithms, then test and compare the control performance among them using either the simulated disturbances or the collected disturbances.

\subsection{Training Results}


%


\begin{figure}[t]
	\centering
	\includegraphics[width=\linewidth]{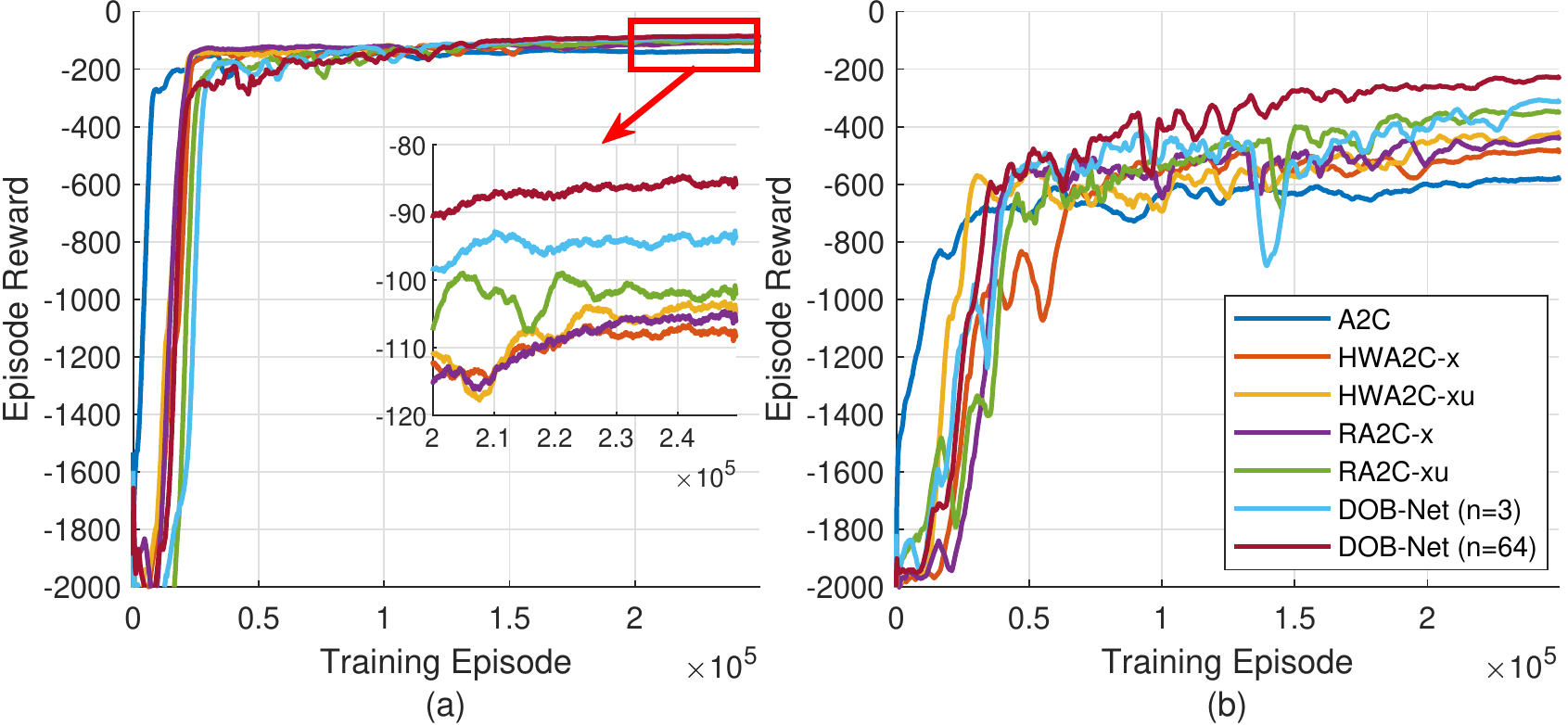}
	\caption{Training rewards, (a) small simulated disturbances (amplitude between 100-120\% of control limits); (b) large simulated disturbances (amplitude between 130-150\% of control limits).}
	\label{fig:training_reward}
\end{figure}

\begin{figure}[b]
	\centering
	\includegraphics[width=\linewidth]{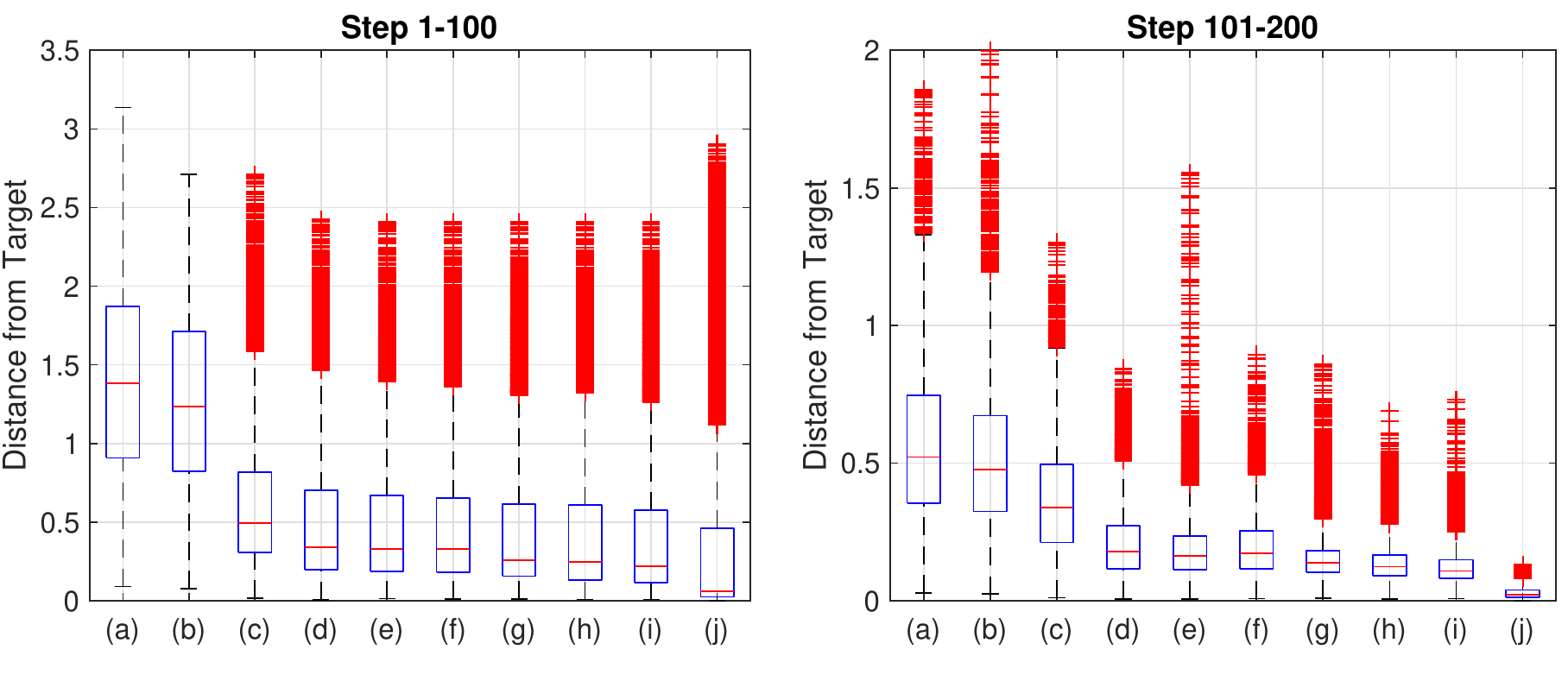}
	\caption{Distribution of distance from target with small simulated disturbances.}
	\label{fig:distance_distribution_simu_small}
\end{figure}



Fig.~\ref{fig:training_reward} proves the performance improvement of RL when considering history information.
When only small disturbances occur, different approaches to use the history information have nearly the same accumulated reward.
While if the disturbances become larger, there is noticeable increase of using additional action input for both HWA2C and RA2C. Also, using RNN instead of history window approach achieves higher accumulated reward, this is reasonable due to the more efficient utilization of the history information. For the DOB-Net algorithms, we notice that using larger embedding size ($n=64$) of disturbance estimate produces higher reward.

\subsection{Test Results on Simulated Disturbances}

\begin{figure}[b]
	\centering
	\includegraphics[width=\linewidth]{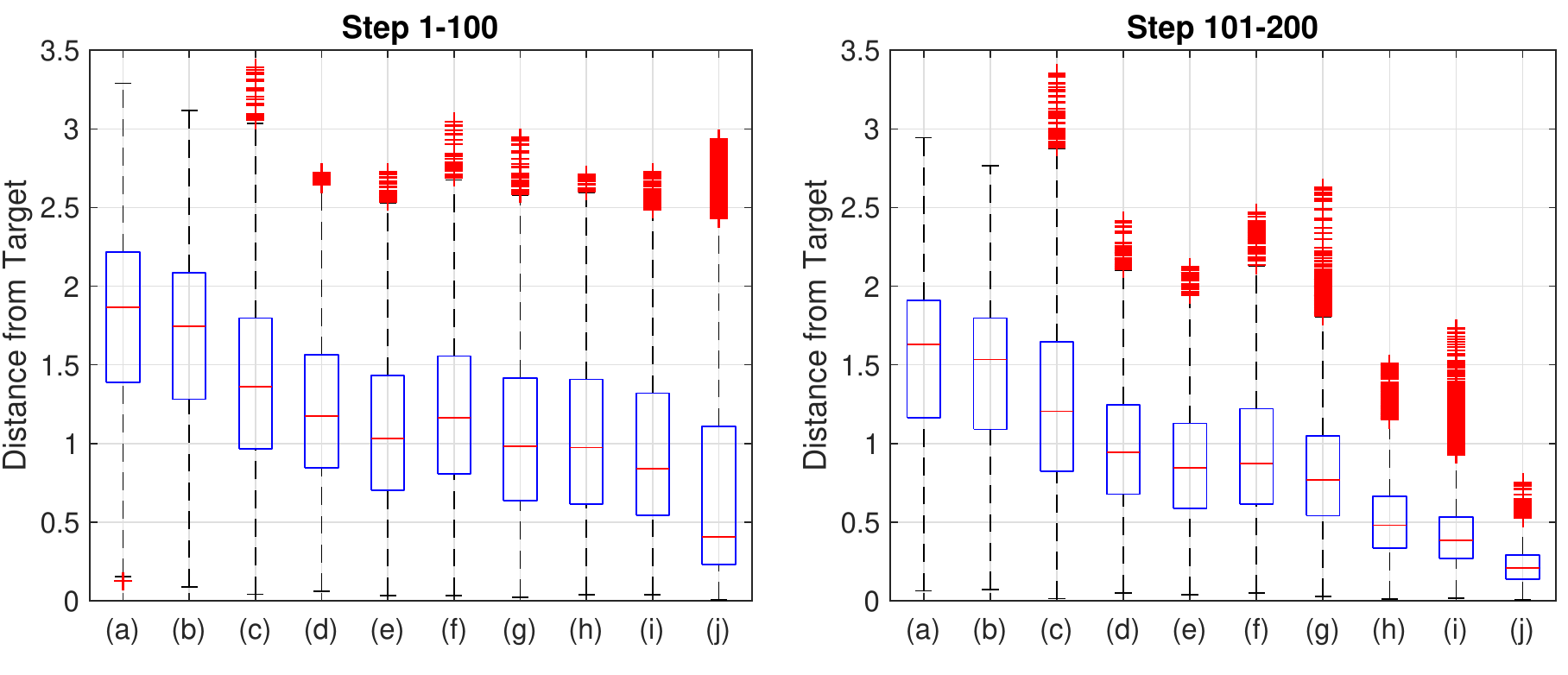}
	\caption{Distribution of distance from target with large simulated disturbances.}
	\label{fig:distance_distribution_simu_large}
\end{figure}

The training reward is not sufficient to compare the performance among different algorithms, we are also interested in state distribution and bounded response (i.e. converged region) of the AUV disturbed by flows. 
As shown in Fig.~\ref{fig:distance_distribution_simu_small} and Fig.~\ref{fig:distance_distribution_simu_large}, the box plot is used to represent and compare the distribution of AUV’s distance from target among different algorithms, in the first half (step 1-100) and second half (step 101-200) of each episode. On each box, the central mark indicates the median, and the bottom and top edges of the box indicate the 25th and 75th percentiles, respectively.
The distance ranges of the second stage are smaller than those of the first stage, since the AUV first observes the disturbance dynamics and then tries to stabilize itself. The results also demonstrate all these algorithms do stabilize the AUV to a certain extend.

\begin{figure*}[ht]
	\begin{minipage}{0.72\textwidth}
		\centering
		\subfigure[RISE]{
			\includegraphics[width=0.3\textwidth]{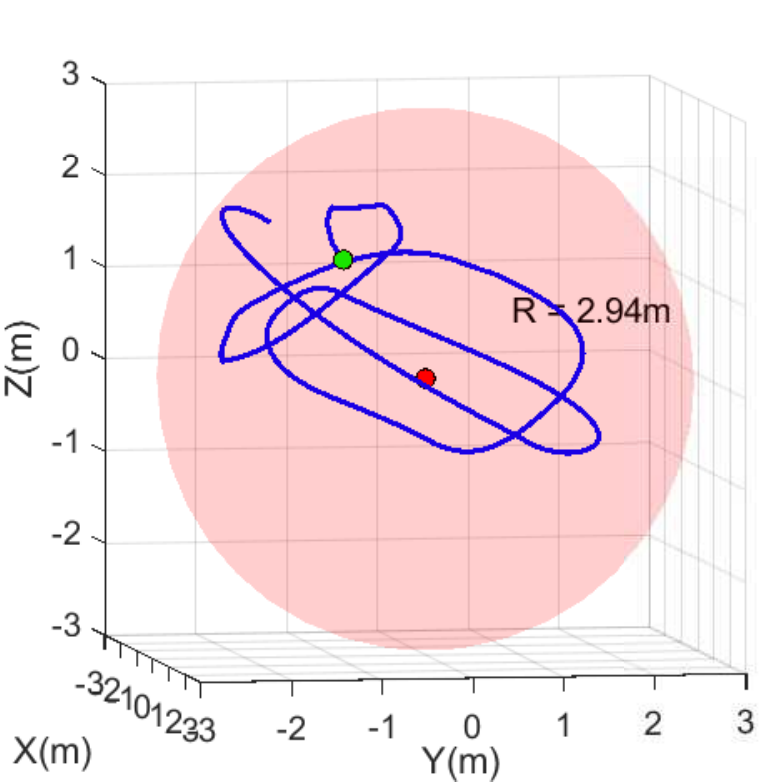}
		}
		\subfigure[DOBC]{
			\includegraphics[width=0.3\textwidth]{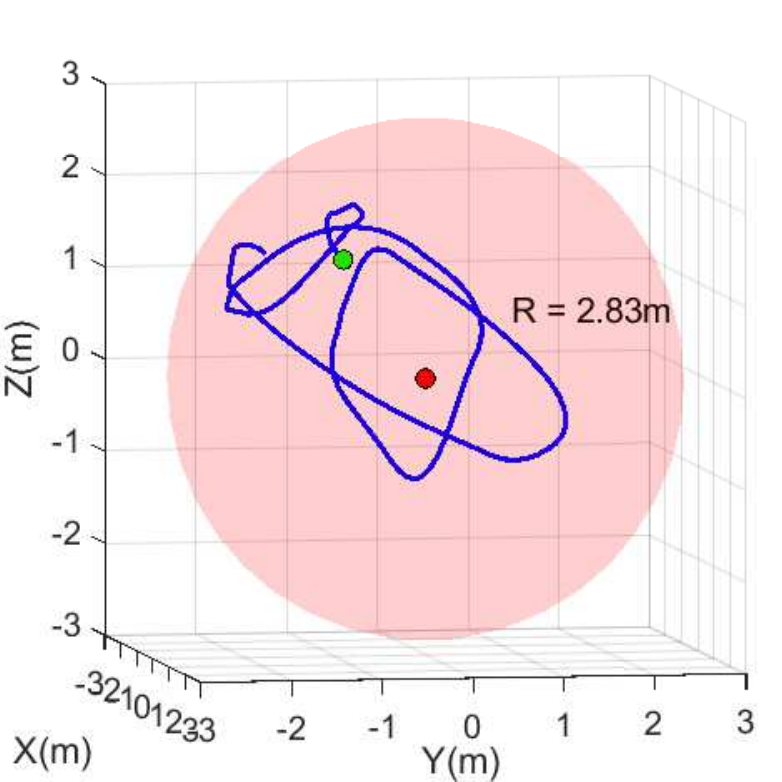}
		}
		\subfigure[A2C]{
			\includegraphics[width=0.3\textwidth]{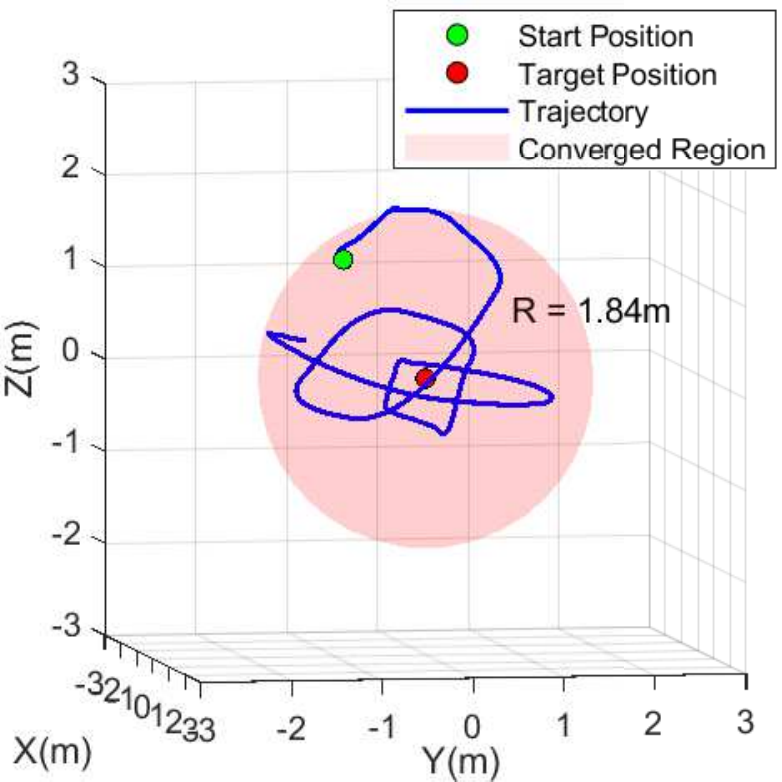}
		}
		
		\subfigure[HWA2C-xu]{
			\includegraphics[width=0.3\textwidth]{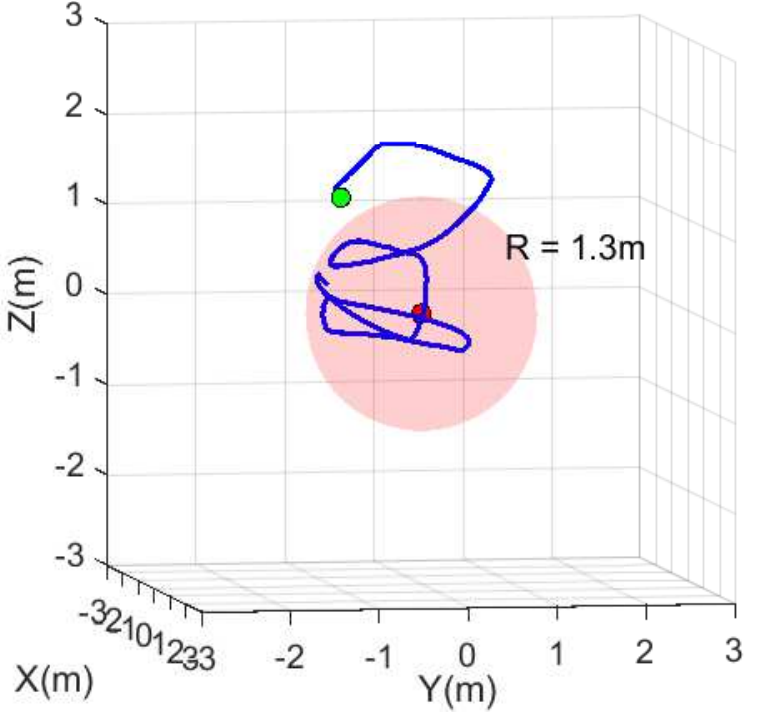}
		}
		\subfigure[RA2C-xu]{
			\includegraphics[width=0.3\textwidth]{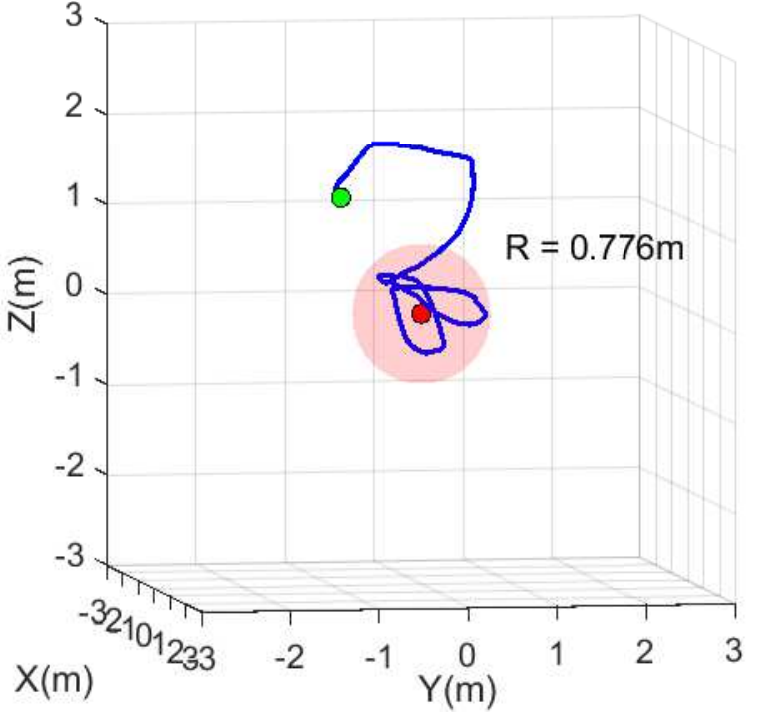}
		}
		\subfigure[DOB-Net]{
			\includegraphics[width=0.3\textwidth]{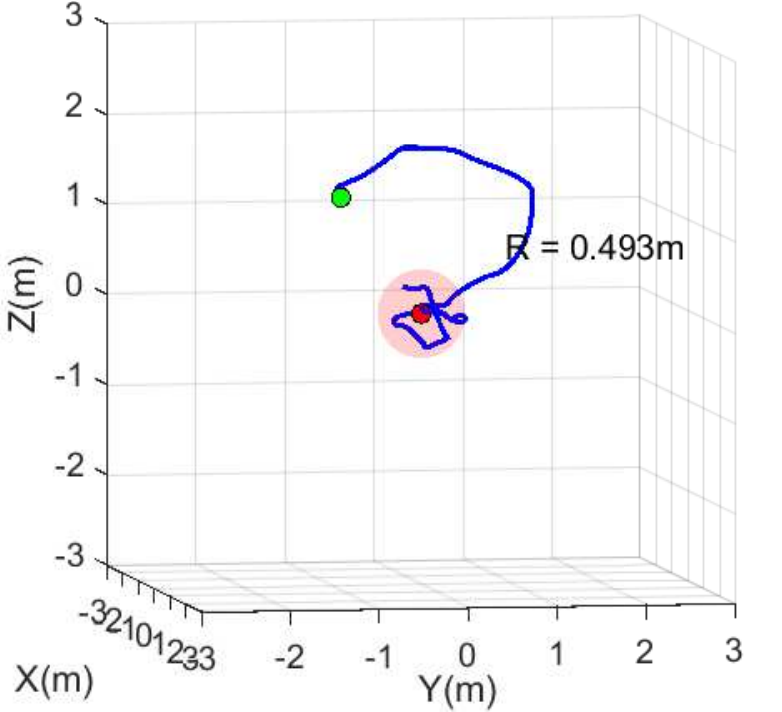}
		}
	\end{minipage}
	\begin{minipage}{0.24\textwidth}
		\includegraphics[width=\linewidth]{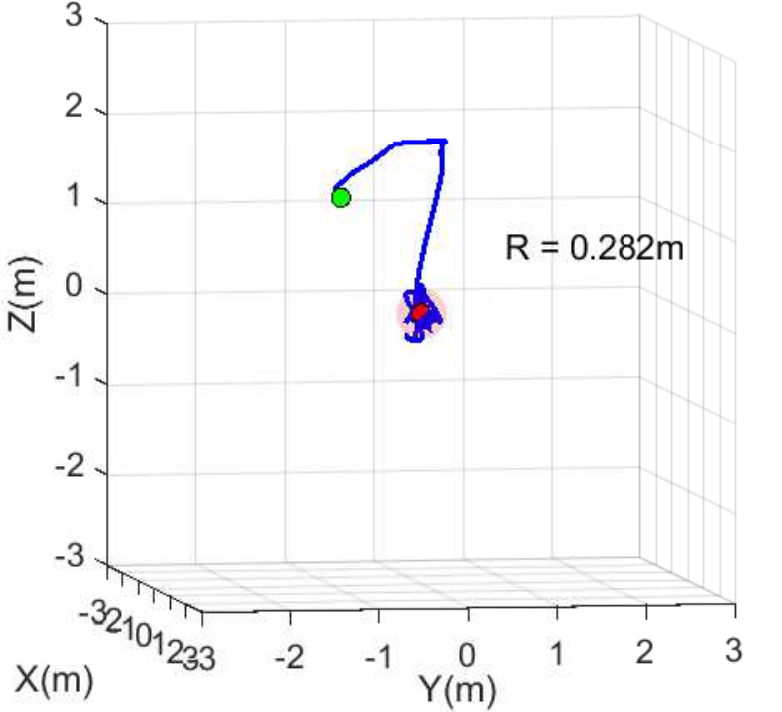}
		\centerline{\small (g) Trajectory Optimization}
	\end{minipage}
	\caption{3D trajectories with large simulated disturbances ($R$ is the radius of the converged region). Note the trajectory optimization assumes the disturbance dynamics is known in advance, thus provides the ideal performance.}
	\label{fig:3d_trajectory_simu_large}
\end{figure*}

Again notice that, the trajectory optimization provides an optimal solution in the case the disturbances through the entire episode are given in advance. Our goal is to narrow the gap between our algorithm and the optimal solution in ideal case. If we focus on the second stage of the episode, it is clear both the history window policies and recurrent policies perform better than the classical RL policy (A2C) and the conventional control schemes (RISE and DOBC), which means the history information does improve the disturbance rejection capability. Also, the recurrent policy considering both past states and actions is even better, proving the RNNs can utilize the history information more efficiently than naively putting together past states and actions into the state space. Furthermore, considering past actions as additional input besides states yields better performance. 

The DOB-Net achieves the best performance among all these algorithms, especially using $n=64$. We believe enlarging the embedding size of the disturbance estimate can provide better representation of the disturbance dynamics. Transforming this embedding from a 64-dimensional variable to a 3-dimensional variable may cause loss of information.
However, even using the best RL algorithms we mentioned so far, the control performance still has a large gap from the trajectory optimization solution. There is still room for further improvements.

\begin{figure}[b]
	\centering
	\includegraphics[width=\linewidth]{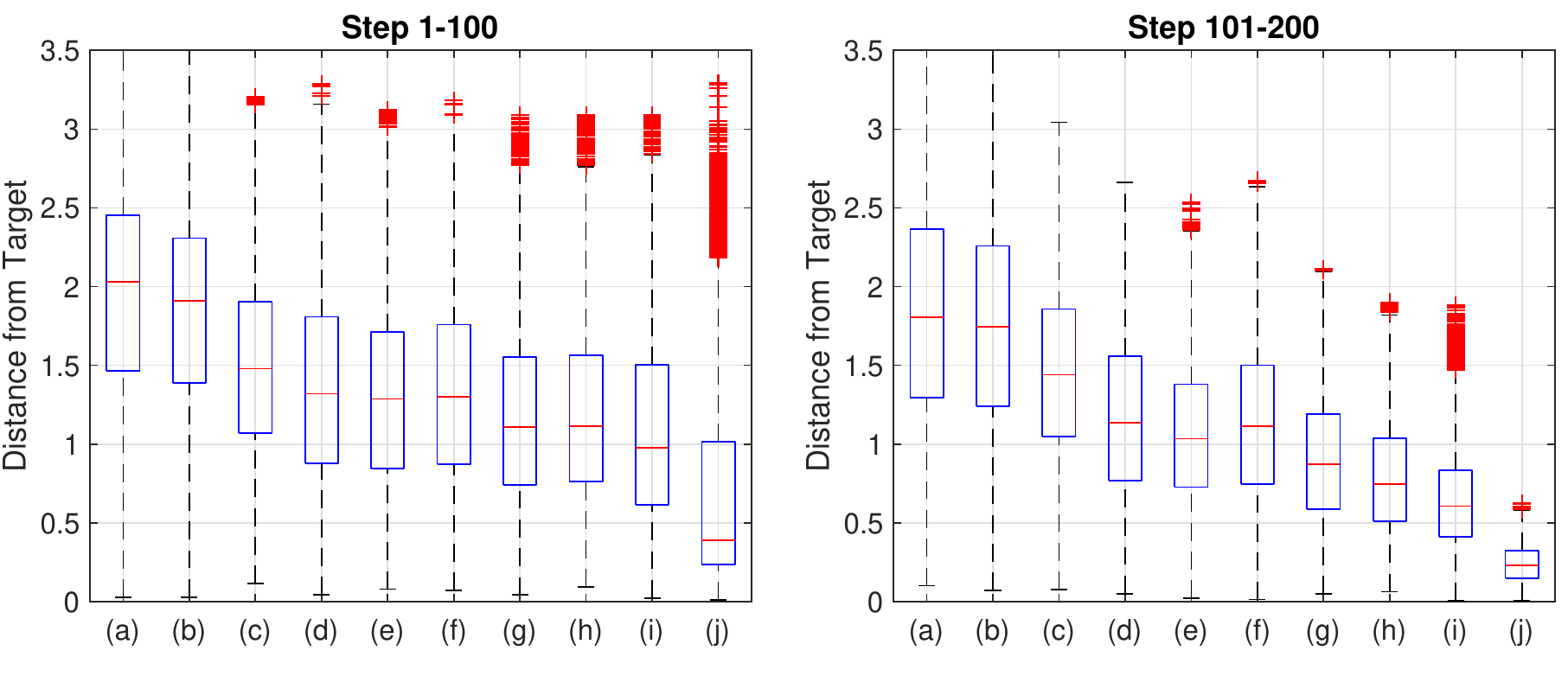}
	\caption{Distribution of distance from target with collected disturbances.}
	\label{fig:distance_distribution_d_real}
\end{figure}

\begin{figure}[b]
	\centering
	\includegraphics[width=0.8\linewidth]{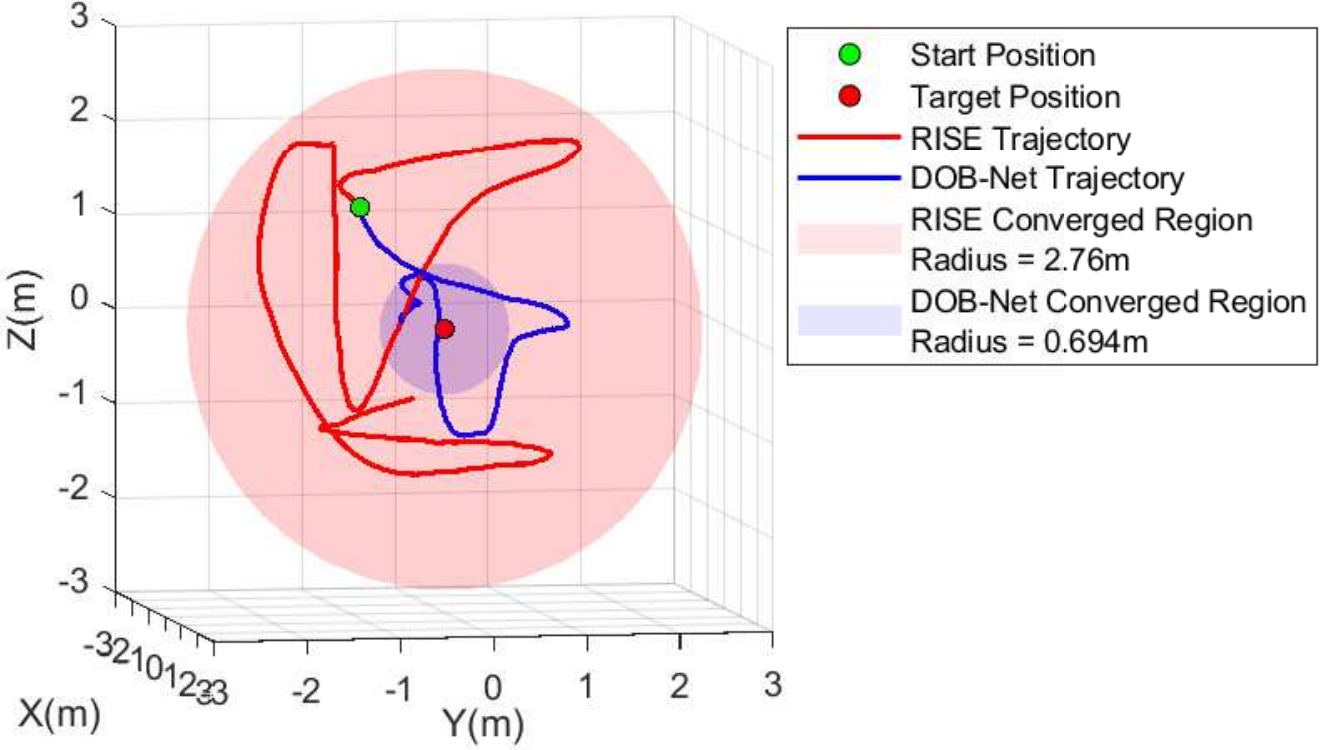}
	\caption{3D trajectories with collected disturbances ($R$ is the radius of the converged region).}
	\label{fig:3d_trajectory_d_real}
\end{figure}

In addition, it is obvious that stronger disturbances lead to worse performance. But we also found larger amplitude range of disturbances gives closer performance between the DOB-Net and the optimal approach (the ratio of medians between the DOB-Net and the trajectory optimization is 502.31\% and 186.65\% respectively for small and large simulated disturbances). This phenomenon might result from the optimal controls for different disturbance patterns with large amplitudes tend to be more similar than those with small amplitudes, thus it is easier for RL to learn a near-optimal policy under larger disturbance amplitudes.

The 3D trajectories of these algorithms subject to large simulated disturbances are compared in Fig.~\ref{fig:3d_trajectory_simu_large}. The red ball represents the AUV's maximum distance from target during last 50 steps, called converged region. According to this region, we can see that the AUV is difficult to achieve satisfactory bounded response using either conventional controller (RISE and DOBC) or classical RL (A2C). While the proposed DOB-Net can significantly narrow the converged region. Using the DOB-Net, the AUV can quickly navigate to the target and stabilize itself within a distance of $0.493 m$ from the target thereafter. However, there is still an obvious gap between the DOB-Net and the optimal trajectory.

\subsection{Test Results on Collected Disturbances}

Besides the simulated disturbances, we also use collected current and wave disturbances (as shown in Fig.~\ref{fig:disturbance} (c)) for testing. Note the collected data is only used for testing, no retraining is required. 
The DOB-Net still has satisfactory performance on real-world disturbance scenarios and outperforms all the other algorithms, which proves the practical effectiveness of the DOB-Net. However, the converged region of all the competitors become larger compared with the simulated case, and the performance between the DOB-Net and the optimal approach gets less closer (the ratio of medians between the DOB-Net and the trajectory optimization increases to 264.88\%, with respect to 186.65\% when using large simulated disturbances). The reason behind is that the collected disturbances are more diverse and complicated, thus have a wider range of amplitudes compared with the simulated case. The proposed algorithms may not be capable of handling such unseen scenarios optimally. This gives rise to another research question, which is to deal with disturbances with a wider range of parameters based on training on small range of parameters. This may require the technique of transfer learning \cite{pan2009survey}.


\section{CONCLUSION \& FUTURE WORK}

This paper proposes an observer-integrated RL approach called DOB-Net, for mobile robot control problems under unknown excessive time-varying disturbances. A disturbance dynamics observer network employing RNNs has been used to imitate and enhance the function of conventional DOB, which produces the embedding of disturbance estimation and prediction. A controller network is designed using the observer outputs as well as current state as inputs, to generate optimal controls. Multiple control and RL algorithms have been tested and compared on position regulation tasks using both simulated disturbances and collected disturbances, the results demonstrate that the proposed DOB-Net does have a significant improvement for the disturbance rejection capacity compared to existing methods.

Currently, the test disturbances are collected in a water tank using wave generator, we plan to seek for the disturbance data from open water environments with natural current and wave for further testing.
Also, we have noticed that the performance of the DOB-Net is worse using the collected disturbances, due to its more complex and diverse dynamics. An interesting future work is to investigate the usage of transfer learning in dealing with real world current and wave disturbances when simulated data and only a small amount of collected data are available.
In addition, the deployment of this method on real-world robotic systems requires future investigation, where the low sample efficiency of generic model-free RL might be a problem. Some model-based approaches are necessary to overcome the constraints of real-time sample collection in the real world.




%

%


\bibliographystyle{./bibliographies/IEEEtran}
\bibliography{./bibliographies/IEEEabrv,./bibliographies/mybibfile}

\end{document}